\documentclass[11pt,a4paper]{article}
\usepackage[hyperref]{emnlp2018}
\usepackage{times}
\usepackage{latexsym}
\usepackage{graphicx} 
\usepackage{overpic}
\usepackage{color}
\usepackage{multirow}
\usepackage{amssymb}
\usepackage{amsmath}
\usepackage{subfigure}
\usepackage{url,comment}
\usepackage{fancyhdr}

\aclfinalcopy % Uncomment this line for the final submission

\title{Interpreting Recurrent and Attention-Based Neural Models: \\a Case  Study on Natural Language Inference}

\author{
	Reza Ghaeini, Xiaoli Z. Fern, Prasad Tadepalli\\
	School of Electrical Engineering and Computer Science, Oregon State University\\
	1148 Kelley Engineering Center, Corvallis, OR 97331-5501, USA\\
	{\tt \{ghaeinim, xfern, tadepall\}@eecs.oregonstate.edu}
	\\}

\date{}

\begin{document}
	\maketitle
	
	\begin{abstract}
Deep learning models have achieved remarkable success in natural language inference (NLI) tasks. While these models are widely explored, they are hard to interpret and it is often unclear how and why they actually work. In this paper, we take a step toward explaining such deep learning based models through a case study on a popular neural model for NLI. In particular, we propose to interpret the intermediate layers of NLI models by visualizing the saliency of attention and LSTM gating signals. We present several examples for which our methods are able to reveal interesting insights and identify the critical information contributing to the model decisions. 
	\end{abstract}

	\section{Introduction}
      Deep learning has achieved tremendous success for many NLP tasks. However, unlike traditional methods that provide optimized weights for human understandable features, the behavior of deep learning models is much harder to interpret. Due to the high dimensionality of word embeddings, and the complex, typically recurrent architectures used for textual data, 
it is often unclear how and why a deep learning model reaches its decisions.
	
There are a few attempts toward explaining/interpreting deep learning-based models, mostly by visualizing the representation of words and/or hidden states, and their importances (via saliency or erasure) on shallow tasks like sentiment analysis and POS tagging \cite{explain_nlp,lrp,erasure,zerosh}. In contrast, we focus on interpreting the gating and attention signals of the intermediate layers of deep models in the challenging task of Natural Language Inference. 
A key concept in explaining deep models is saliency, which determines what is critical for the final decision of a deep model. So far, saliency has only been used to illustrate the impact of word embeddings. In this paper, we extend this concept to the intermediate layer of deep models to examine the saliency of attention as well as the LSTM gating signals to understand the behavior of these components and their impact on the final decision. 

We make two main contributions. First, we introduce new strategies for interpreting the behavior of deep models in their intermediate layers, specifically, by examining the saliency of the attention and the gating signals.	Second, we provide an extensive analysis of the state-of-the-art model for the NLI task and show that our methods reveal interesting insights not available from traditional methods of inspecting attention and word saliency.

In this paper, our focus was on NLI, which is a fundamental NLP task that requires both understanding and reasoning. Furthermore, the state-of-the-art NLI models employ complex neural architectures involving key mechanisms, such as attention and repeated reading, widely seen in successful models for other NLP tasks. As such, we expect our methods to be potentially useful for other natural understanding tasks as well. 
	
    \begin{figure*}[ht]
		\centering
		\includegraphics[width=.9\textwidth]{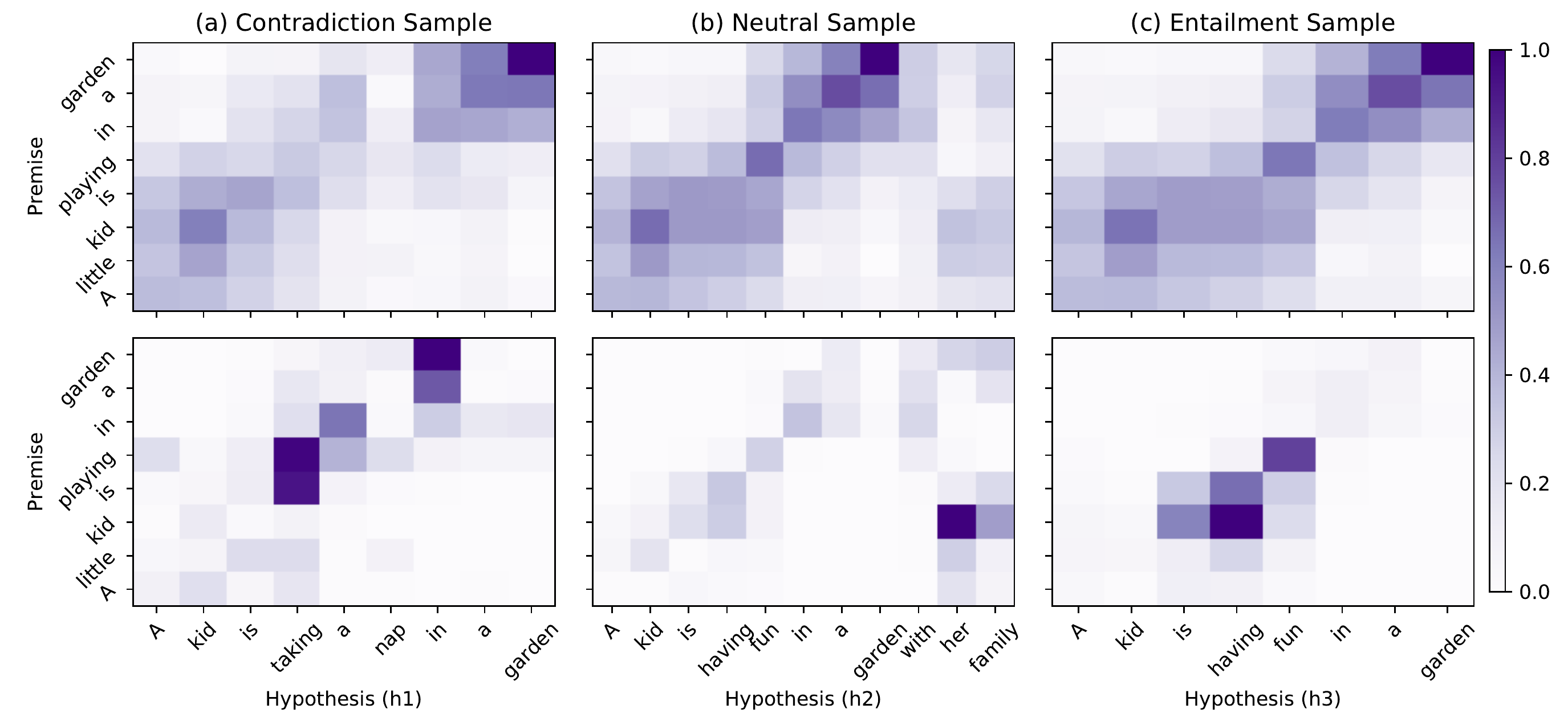}
		\caption{Normalized attention and attention saliency visualization. Each column shows visualization of one sample. Top plots depict attention visualization and bottom ones represent attention saliency visualization. Predicted (the same as Gold) label of each sample is shown on top of each column. \label{fig:att}}
%        \vspace{-0.15in}
	\end{figure*}
    
\section{Task and Model}
%\vspace{-0.05in}
In NLI \cite{snli}, we are given two sentences, a premise and a hypothesis, the goal is to decide the logical relationship (\emph{Entailment}, \emph{Neutral}, or \emph{Contradiction}) between them. 
    
Many of the top performing NLI models \cite{drbilstm,cafe,elmo,socher,gong2017,ibm2017,him2017}, are variants of the ESIM model \cite{him2017}, which we choose to analyze in this paper. 
ESIM reads the sentences independently using LSTM at first, and then applies attention to align/contrast the sentences. Another round of LSTM reading then produces the final representations, which are compared to make the prediction. Detailed description of ESIM can be found in the Appendix. 

Using the SNLI \cite{snli} data, we train two variants of ESIM, with dimensionality 50 and 300 respectively, referred to as ESIM-50 and ESIM-300 in the remainder of the paper.

\section{Visualization of Attention and Gating}
%\vspace{-0.05in}
In this work, we are primarily interested in the internal workings of the NLI model. In particular, we focus on the attention and the gating signals of LSTM readers, and how they contribute to the decisions of the model.
    
\subsection{Attention}
Attention has been widely used in many NLP tasks \cite{dgr, ga-reader, nmt} and is probably one of the most critical parts that affects the inference decisions. Several 
pieces of prior work in NLI have attempted to visualize the attention layer 
to provide some understanding of their models \cite{drbilstm,google2016}. Such visualizations generate a heatmap representing the similarity between the 
hidden states of the premise and the hypothesis  (Eq.~\ref{eq:energy} of Appendix). Unfortunately the similarities are often the same 
regardless of the decision.

Let us consider the following example, where the same premise \textit{``A kid is playing in the garden''}, is paired with three different hypotheses:
\begin{enumerate}
\vspace{-0.1in}
\item [h1:] \textit{A kid is taking a nap in the garden}
\vspace{-0.1in}
\item [h2:] \textit{A kid is having fun in the garden with her family}
\vspace{-0.1in}
\item [h3:] \textit{A kid is having fun in the garden}
\end{enumerate}
\vspace{-0.1in} 
Note that the ground truth relationships are Contradiction, Neutral, and Entailment, respectively. 

The first row of Fig.~\ref{fig:att} shows the visualization of normalized attention for the three cases produced by ESIM-50, which makes correct predictions for all of them. As we can see from the figure, the three attention maps are fairly similar despite the completely different decisions. The key issue is that the attention visualization only allows us to see how the model aligns the premise with the hypothesis, but does not show \textit{how such alignment impacts the decision.} This prompts us to consider the saliency of attention.

\begin{figure*}[ht]
	\centering
	\includegraphics[width=.95\textwidth]{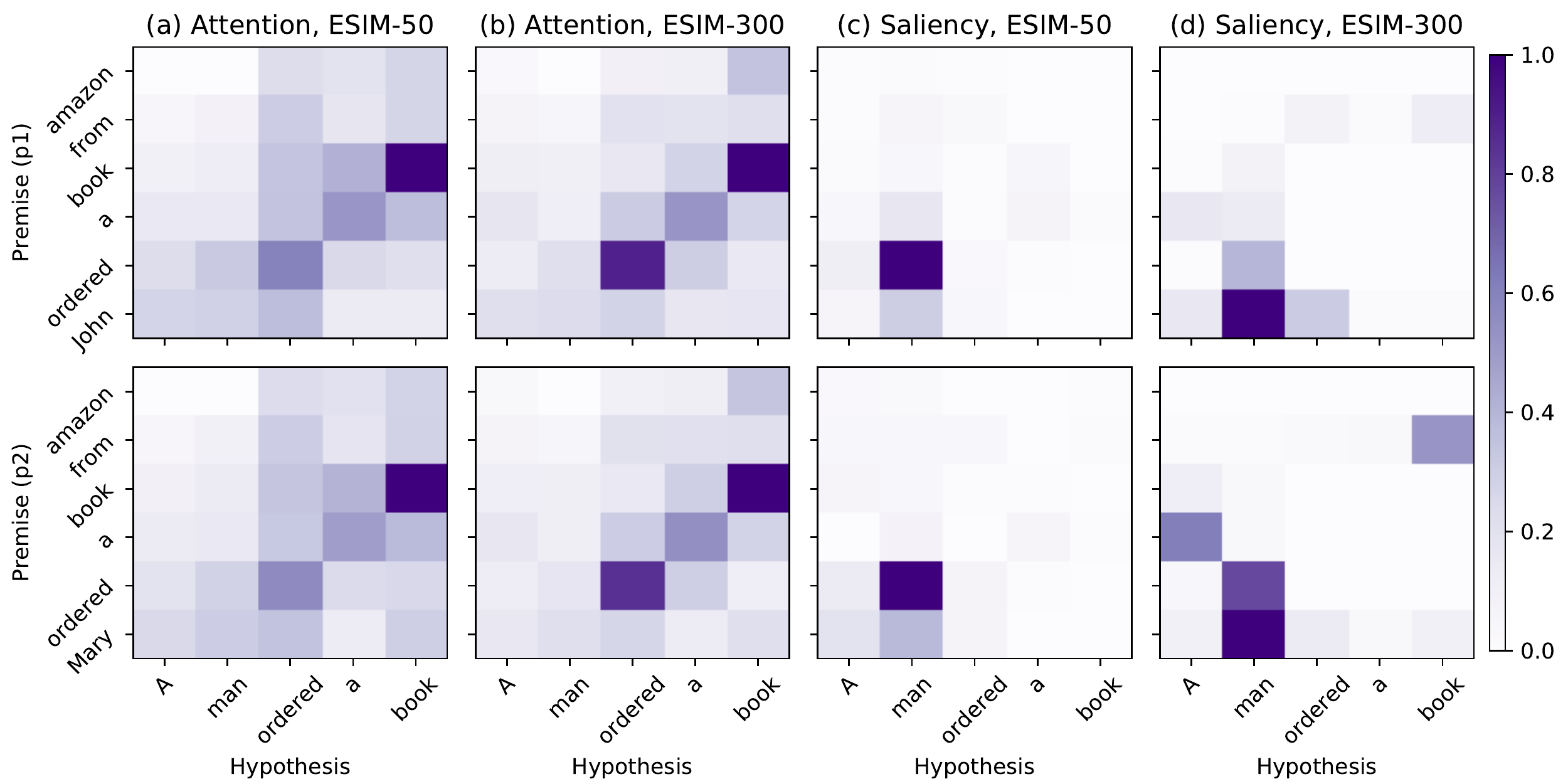}
	\caption{Normalized attention and attention saliency visualizations of two examples (p1 and p2) for ESIM-50 (a) and ESIM-300 (b) models. Each column indicates visualization of a model and each row represents visualization of one example. \label{fig:att_saliency_gender}}
	% \vspace{-0.1in}
\end{figure*}

\subsubsection{Attention Saliency}
The concept of saliency was first introduced in vision for visualizing the spatial support on an image for a particular object class \cite{vision_insp}. In NLP, saliency has been used to study the importance of words toward a final decision \cite{explain_nlp} . 

We propose to examine the saliency of attention. Specifically, given a premise-hypothesis pair and the model's decision $y$, we consider the similarity between a pair of premise and hypothesis hidden states $e_{ij}$ as a variable. The score of the decision $S(y)$ is thus a function of $e_{ij}$ for all $i$ and $j$. The saliency of $e_{ij}$ is then defined to be $|\frac{\partial S(y)}{\partial{e_{ij}}}|$.

The second row of Fig.~\ref{fig:att} presents the attention saliency map for the three examples acquired by the same ESIM-50 model.  Interestingly, the saliencies are clearly different across the examples, each highlighting different parts of the alignment. Specifically, for h1, we see the alignment between ``is playing'' and ``taking a nap'' and the alignment of ``in a garden'' to have the most prominent saliency toward the decision of Contradiction. For h2, the alignment of ``kid'' and ``her family'' seems to be the most salient for the decision of Neutral.  Finally, for h3, the alignment between ``is having fun'' and ``kid is playing'' have the strongest impact toward the decision of Entailment. 

From this example, we can see that by inspecting the attention saliency, we effectively pinpoint which part of the alignments contribute most critically to the final prediction whereas simply visualizing the attention itself reveals little information.

\subsubsection{Comparing Models}
In the previous examples, we study the behavior of the same model on different inputs. Now we use the attention saliency to compare the two different ESIM models: ESIM-50 and ESIM-300.

Consider two examples with a shared hypothesis of \textit{``A man ordered a book''} and premise:
\begin{enumerate}
\vspace{-0.1in}
\item [p1:] \textit{John ordered a book from amazon}
\vspace{-0.1in}
\item [p2:] \textit{Mary ordered a book from amazon}
\end{enumerate}
\vspace{-0.1in}
Here ESIM-50 fails to capture the gender connections of the two different names and predicts Neutral for both inputs, whereas ESIM-300 correctly predicts Entailment for the first case and Contradiction for the second.

In the first two columns of Fig.~\ref{fig:att_saliency_gender} (column a and b) we visualize the attention of the two examples for ESIM-50 (left) and ESIM-300 (right) respectively. Although the two models make different predictions, their attention maps appear qualitatively similar.  
    
In contrast, columns 3-4 of Fig.~\ref{fig:att_saliency_gender} (column c and d) present the attention saliency for the two examples by ESIM-50 and ESIM-300 respectively. We see that for both examples,  ESIM-50 primarily focused on the alignment of ``ordered'', whereas ESIM-300 focused more on the alignment of ``John'' and ``Mary'' with ``man''. 
It is interesting to note that ESIM-300 does not appear to learn significantly different similarity values compared to ESIM-50 for the two critical pairs of words (``John'', ``man'') and (``Mary'', ``man'') based on the attention map. The saliency map, however, reveals that the two models use these values quite differently, with 
only ESIM-300 correctly focusing on them. 
   
\begin{figure*}[ht]
		\centering
        \begin{overpic}[width=\textwidth]{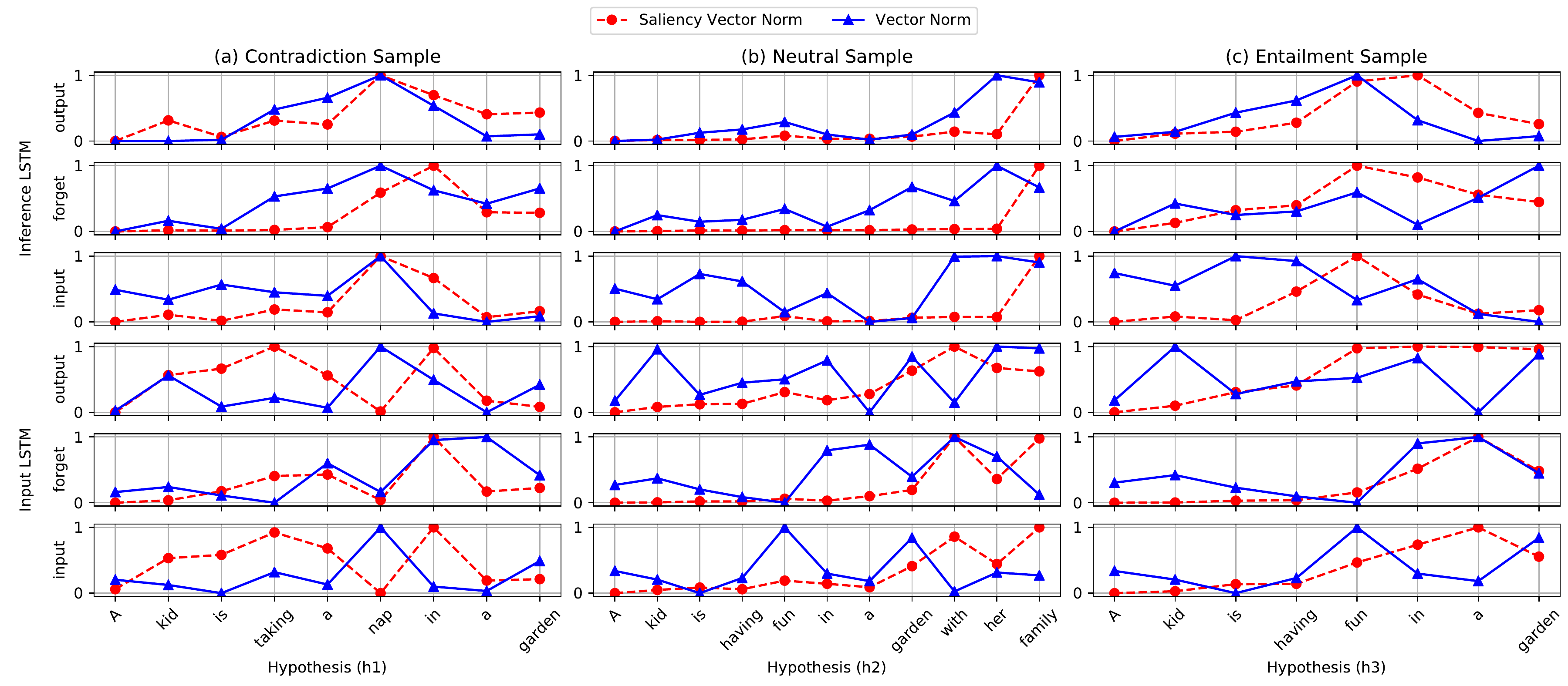}
        \put(0,22.3){\color{black}\rule{\textwidth}{1pt}}
        \put(2.75,5.2){\color{black}\line(0,1){34.1}}
        \end{overpic}
		\caption{Normalized signal and saliency norms for the input and inference LSTMs (forward) of ESIM-50 for three examples. The bottom (top) three rows show the signals of the input (inference) LSTM. Each row shows one of the three gates (input, forget and output). }
\label{fig:f_gate_sig}
%\vspace{-0.1in}
\end{figure*}
    
	\subsection{LSTM Gating Signals}
LSTM gating signals determine the flow of information. In other words, they indicate how LSTM reads the word sequences and how the information from different parts is captured and combined. LSTM gating signals are rarely analyzed, possibly due to their high dimensionality and complexity. In this work, we consider both the gating signals and their saliency, which is computed as the partial derivative of the score of the final decision with respect to each gating signal. 

Instead of considering individual dimensions of the gating signals, we aggregate them to consider their norm, both for the signal and for its saliency. Note that ESIM models have two LSTM layers, the first (input) LSTM performs the input encoding and the second (inference) LSTM generates the representation for inference.

In Fig.~\ref{fig:f_gate_sig} we plot the normalized signal and saliency norms for different gates (input, forget, output)\footnote{We also examined the memory cell but it shows very similar behavior with the output gate and is hence omitted.} of the Forward input (bottom three rows) and inference (top three rows) LSTMs. These results are produced by the ESIM-50 model for the three examples of Section 3.1, one for each column.

From the figure, we first note that the saliency tends to be somewhat consistent across different gates within the same LSTM, suggesting that we can interpret them jointly to identify parts of the sentence important for the model's prediction.

Comparing across examples, we see that the saliency curves show pronounced differences across the examples. For instance, the saliency pattern of the Neutral example is significantly different from the other two examples, and heavily concentrated toward the end of the sentence (``with her family''). Note that without this part of the sentence, the relationship would have been Entailment. The focus (evidenced by its strong saliency and strong gating signal) on this particular part, which presents information not available from the premise, explains the model's decision of Neutral. 

Comparing the behavior of the input LSTM and the inference LSTM, we observe interesting shifts of focus. In particular, we see that the inference LSTM tends to see much more concentrated saliency over key parts of the sentence, whereas the input LSTM sees more spread of saliency. For example, for the Contradiction example, the input LSTM sees high saliency for both ``taking'' and ``in'', whereas the inference LSTM primarily focuses on ``nap'', which is the key word suggesting a Contradiction. 
Note that ESIM uses attention between the input and inference LSTM layers to align/contrast the sentences, hence it makes sense that the inference LSTM is more focused on the critical differences between the sentences. This is also observed for the Neutral example as well.

It is worth noting that, while revealing similar general trends, the backward LSTM can sometimes focus on different parts of the sentence (e.g., see Fig.~\ref{fig:b_gate_sig} of Appendix), suggesting the forward and backward readings provide complementary understanding of the sentence.
    
\section{Conclusion}
%\vspace{-.05in}
We propose new visualization and interpretation strategies for neural models to understand how and why they work. We demonstrate the effectiveness of the proposed strategies on a complex task (NLI). Our strategies are able to provide interesting insights not achievable by previous explanation techniques. Our future work will extend our study to consider other NLP tasks and models with the goal of producing useful insights for further improving these models.
\bibliography{emnlp2018}

\begin{thebibliography}{17}
\expandafter\ifx\csname natexlab\endcsname\relax\def\natexlab#1{#1}\fi

\bibitem[{Arras et~al.(2017)Arras, Montavon, M{\"{u}}ller, and Samek}]{lrp}
Leila Arras, Gr{\'{e}}goire Montavon, Klaus{-}Robert M{\"{u}}ller, and Wojciech
  Samek. 2017.
\newblock Explaining recurrent neural network predictions in sentiment
  analysis.
\newblock In \emph{Proceedings of the 8th Workshop on Computational Approaches
  to Subjectivity, Sentiment and Social Media Analysis, WASSA@EMNLP 2017,
  Copenhagen, Denmark, September 8, 2017}, pages 159--168.

\bibitem[{Bahdanau et~al.(2014)Bahdanau, Cho, and Bengio}]{nmt}
Dzmitry Bahdanau, Kyunghyun Cho, and Yoshua Bengio. 2014.
\newblock Neural machine translation by jointly learning to align and
  translate.
\newblock \emph{CoRR}, abs/1409.0473.

\bibitem[{Bowman et~al.(2015)Bowman, Angeli, Potts, and Manning}]{snli}
Samuel~R. Bowman, Gabor Angeli, Christopher Potts, and Christopher~D. Manning.
  2015.
\newblock A large annotated corpus for learning natural language inference.
\newblock In \emph{Proceedings of the 2015 Conference on Empirical Methods in
  Natural Language Processing, {EMNLP} 2015, Lisbon, Portugal, September 17-21,
  2015}, pages 632--642.

\bibitem[{Chen et~al.(2017)Chen, Zhu, Ling, Wei, Jiang, and Inkpen}]{him2017}
Qian Chen, Xiaodan Zhu, Zhen{-}Hua Ling, Si~Wei, Hui Jiang, and Diana Inkpen.
  2017.
\newblock Enhanced {LSTM} for natural language inference.
\newblock In \emph{Proceedings of the 55th Annual Meeting of the Association
  for Computational Linguistics, {ACL} 2017, Vancouver, Canada, July 30 -
  August 4, Volume 1: Long Papers}, pages 1657--1668.

\bibitem[{Dhingra et~al.(2017)Dhingra, Liu, Yang, Cohen, and
  Salakhutdinov}]{ga-reader}
Bhuwan Dhingra, Hanxiao Liu, Zhilin Yang, William~W. Cohen, and Ruslan
  Salakhutdinov. 2017.
\newblock Gated-attention readers for text comprehension.
\newblock In \emph{Proceedings of the 55th Annual Meeting of the Association
  for Computational Linguistics, {ACL} 2017, Vancouver, Canada, July 30 -
  August 4, Volume 1: Long Papers}, pages 1832--1846.

\bibitem[{Ghaeini et~al.(2018{\natexlab{a}})Ghaeini, Fern, Shahbazi, and
  Tadepalli}]{dgr}
Reza Ghaeini, Xiaoli~Z. Fern, Hamed Shahbazi, and Prasad Tadepalli.
  2018{\natexlab{a}}.
\newblock Dependent gated reading for cloze-style question answering.
\newblock \emph{CoRR}, abs/1805.10528.

\bibitem[{Ghaeini et~al.(2018{\natexlab{b}})Ghaeini, Hasan, Datla, Liu, Lee,
  Qadir, Ling, Prakash, Fern, and Farri}]{drbilstm}
Reza Ghaeini, Sadid~A. Hasan, Vivek~V. Datla, Joey Liu, Kathy Lee, Ashequl
  Qadir, Yuan Ling, Aaditya Prakash, Xiaoli~Z. Fern, and Oladimeji Farri.
  2018{\natexlab{b}}.
\newblock Dr-bilstm: Dependent reading bidirectional {LSTM} for natural
  language inference.
\newblock \emph{{NAACL} {HLT} 2018, The 2018 Conference of the North American
  Chapter of the Association for Computational Linguistics: Human Language
  Technologies}.

\bibitem[{Gong et~al.(2017)Gong, Luo, and Zhang}]{gong2017}
Yichen Gong, Heng Luo, and Jian Zhang. 2017.
\newblock Natural language inference over interaction space.
\newblock \emph{CoRR}, abs/1709.04348.

\bibitem[{Li et~al.(2016)Li, Chen, Hovy, and Jurafsky}]{explain_nlp}
Jiwei Li, Xinlei Chen, Eduard~H. Hovy, and Dan Jurafsky. 2016.
\newblock Visualizing and understanding neural models in {NLP}.
\newblock In \emph{{NAACL} {HLT} 2016, The 2016 Conference of the North
  American Chapter of the Association for Computational Linguistics: Human
  Language Technologies, San Diego California, USA, June 12-17, 2016}, pages
  681--691.

\bibitem[{Li et~al.(2017)Li, Monroe, and Jurafsky}]{erasure}
Jiwei Li, Will Monroe, and Dan Jurafsky. 2017.
\newblock Understanding neural networks through representation erasure.
\newblock \emph{CoRR}, abs/1612.08220.

\bibitem[{McCann et~al.(2017)McCann, Bradbury, Xiong, and Socher}]{socher}
Bryan McCann, James Bradbury, Caiming Xiong, and Richard Socher. 2017.
\newblock Learned in translation: Contextualized word vectors.
\newblock In \emph{Advances in Neural Information Processing Systems 30: Annual
  Conference on Neural Information Processing Systems 2017, 4-9 December 2017,
  Long Beach, CA, {USA}}, pages 6297--6308.

\bibitem[{Parikh et~al.(2016)Parikh, T{\"{a}}ckstr{\"{o}}m, Das, and
  Uszkoreit}]{google2016}
Ankur~P. Parikh, Oscar T{\"{a}}ckstr{\"{o}}m, Dipanjan Das, and Jakob
  Uszkoreit. 2016.
\newblock A decomposable attention model for natural language inference.
\newblock In \emph{Proceedings of the 2016 Conference on Empirical Methods in
  Natural Language Processing, {EMNLP} 2016, Austin, Texas, USA, November 1-4,
  2016}, pages 2249--2255.

\bibitem[{Peters et~al.(2018)Peters, Neumann, Iyyer, Gardner, Clark, Lee, and
  Zettlemoyer}]{elmo}
Matthew~E. Peters, Mark Neumann, Mohit Iyyer, Matt Gardner, Christopher Clark,
  Kenton Lee, and Luke Zettlemoyer. 2018.
\newblock Deep contextualized word representations.
\newblock \emph{CoRR}, abs/1802.05365.

\bibitem[{Rei and S{\o}gaard(2018)}]{zerosh}
Marek Rei and Anders S{\o}gaard. 2018.
\newblock Zero-shot sequence labeling: Transferring knowledge from sentences to
  tokens.
\newblock In \emph{Proceedings of the 2018 Conference of the North American
  Chapter of the Association for Computational Linguistics: Human Language
  Technologies, {NAACL-HLT} 2018, New Orleans, Louisiana, USA, June 1-6, 2018,
  Volume 1 (Long Papers)}, pages 293--302.

\bibitem[{Simonyan et~al.(2013)Simonyan, Vedaldi, and Zisserman}]{vision_insp}
Karen Simonyan, Andrea Vedaldi, and Andrew Zisserman. 2013.
\newblock Deep inside convolutional networks: Visualising image classification
  models and saliency maps.
\newblock \emph{CoRR}, abs/1312.6034.

\bibitem[{Tay et~al.(2018)Tay, Tuan, and Hui}]{cafe}
Yi~Tay, Luu~Anh Tuan, and Siu~Cheung Hui. 2018.
\newblock A compare-propagate architecture with alignment factorization for
  natural language inference.
\newblock \emph{CoRR}, abs/1801.00102.

\bibitem[{Wang et~al.(2017)Wang, Hamza, and Florian}]{ibm2017}
Zhiguo Wang, Wael Hamza, and Radu Florian. 2017.
\newblock Bilateral multi-perspective matching for natural language sentences.
\newblock In \emph{Proceedings of the Twenty-Sixth International Joint
  Conference on Artificial Intelligence, {IJCAI} 2017, Melbourne, Australia,
  August 19-25, 2017}, pages 4144--4150.

\end{thebibliography}
\bibliographystyle{acl_natbib_nourl}

\appendix

\section{Model}

\begin{figure}[ht]
	\centering
	\includegraphics[width=.4\textwidth]{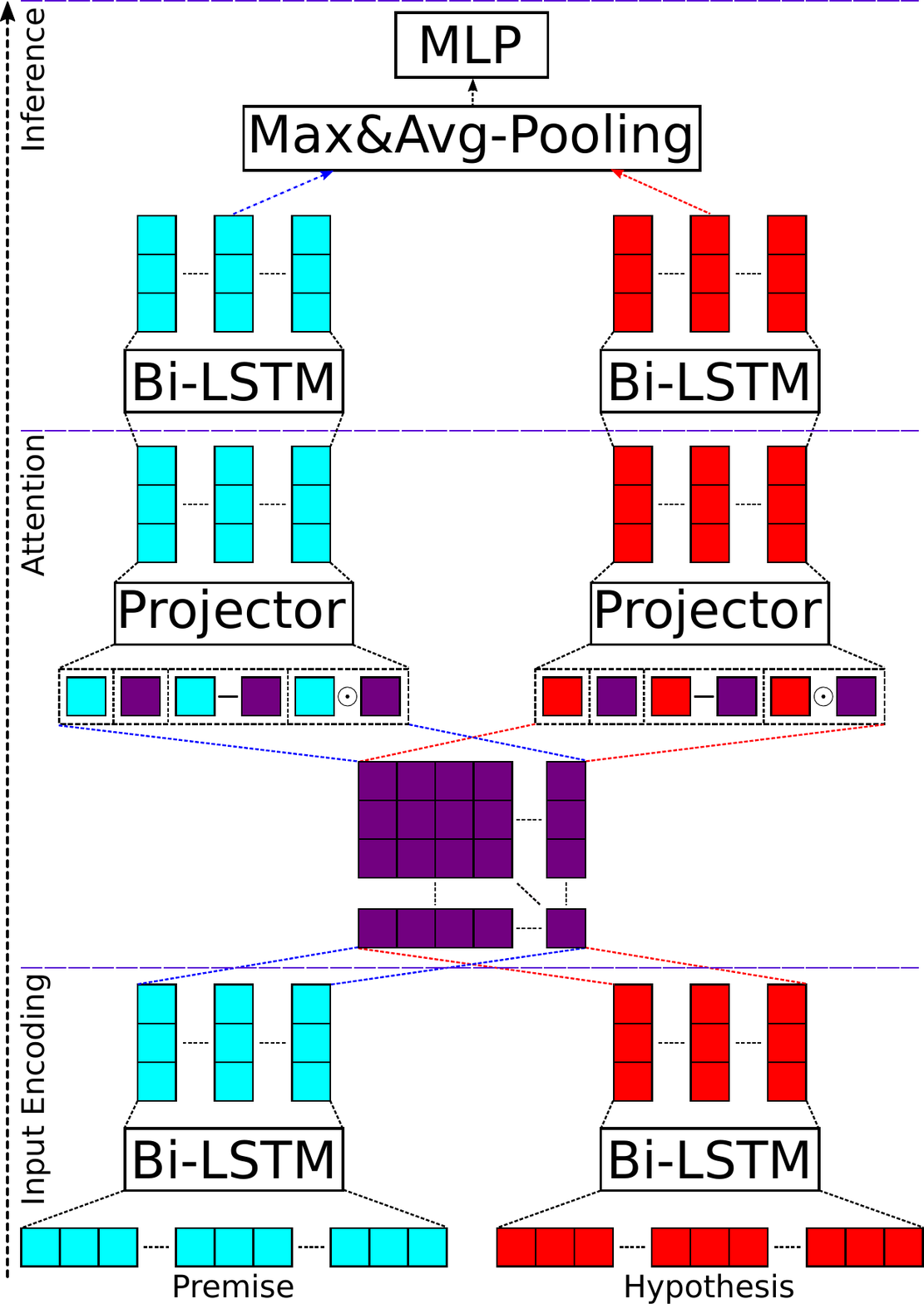}
	\caption{A high-level view of ESIM model.\label{fig:model}}
\end{figure}

\noindent In this section we describe the ESIM model. We divide ESIM to three main parts: 1) input encoding, 2) attention, and 3) inference. Figure~\ref{fig:model} demonstrates a high-level view of the ESIM framework. 

Let $u=[u_1, \cdots, u_n]$ and $v=[v_1, \cdots, v_m]$ be the given premise with length $n$ and hypothesis with length $m$ respectively, where $u_i, v_j \in \mathbb{R}^r$ are word embeddings of $r$-dimensional vector. The goal is to predict a label $y$ that indicates the logical relationship between premise $u$ and hypothesis $v$. Below we briefly explain the aforementioned parts.

\subsection{Input Encoding}
\label{sec:enc}

It utilizes a bidirectional LSTM (BiLSTM) for encoding the given premise and hypothesis using Equations \ref{eq:enc:pout} and \ref{eq:enc:hout} respectively.

\begin{equation}
\hat{u} = \textit{BiLSTM}(u) 
\label{eq:enc:pout}
\end{equation}

\begin{equation}
\hat{v} = \textit{BiLSTM}(v) 
\label{eq:enc:hout}
\end{equation}

\noindent where $\hat{u} \in \mathbb{R}^{n \times 2d}$ and $\hat{v} \in \mathbb{R}^{m \times 2d}$ are the reading sequences of $u$ and $v$ respectively. 

\subsection{Attention}
\label{sec:att}

\noindent It employs a soft alignment method to associate the relevant sub-components between the given premise and hypothesis. Equation~\ref{eq:energy} (energy function) computes the unnormalized attention weights as the similarity of hidden states of the premise and hypothesis.

\begin{equation}
e_{ij} = \hat{u}_i \hat{v}_j^T,  \quad  i \in [1,n], j \in [1,m]
\label{eq:energy}
\end{equation}

\noindent where $\hat{u}_i$ and $\hat{v}_j$ are the hidden representations of $u$ and $v$ respectively which are computed earlier in Equations \ref{eq:enc:pout} and \ref{eq:enc:hout}. Next, for each word in either premise or hypothesis, the relevant semantics in the other sentence is extracted and composed according to $e_{ij}$. Equations \ref{eq:att:p} and \ref{eq:att:h} provide formal and specific details of this procedure.

\begin{equation}
\tilde{u}_i = \sum_{j=1}^{m} \frac{\exp(e_{ij})}{\sum_{k=1}^{m} \exp(e_{ik})} \hat{v}_j, \quad i \in [1,n]
\label{eq:att:p}
\end{equation}
\begin{equation}
\tilde{v}_j = \sum_{i=1}^{n} \frac{\exp(e_{ij})}{\sum_{k=1}^{n} \exp(e_{kj})} \hat{u}_i, \quad j \in [1,m]
\label{eq:att:h}
\end{equation}

\noindent where $\tilde{u}_i$ represents the extracted relevant information of $\hat{v}$ by attending to $\hat{u}_i$ while $\tilde{v}_j$ represents the extracted relevant information of $\hat{u}$ by attending to $\hat{v}_j$. Next, it passes the enriched information through a projector layer which produce the final output of attention stage. Equations \ref{eq:prj:p} and \ref{eq:prj:h} formally represent this process.

\begin{equation}
\begin{split}
a_i &= [\hat{u}_i, \tilde{u}_i, \hat{u}_i -\tilde{u}_i, \hat{u}_i \odot \tilde{u}_i] \\
p_i &=  \textit{ReLU}(W_p a_i + b_p)
\end{split}
\label{eq:prj:p}
\end{equation}
\begin{equation}
\begin{split}
b_j &= [\hat{v}_j, \tilde{v}_j, \hat{v}_j -\tilde{v}_j, \hat{v}_j \odot \tilde{v}_j] \\
q_j &=  \textit{ReLU}(W_p b_j + b_p)
\end{split}
\label{eq:prj:h}
\end{equation}

\noindent Here $\odot$ stands for element-wise product while $W_p \in \mathbb{R}^{8d\times d}$ and $b_p \in \mathbb{R}^{d}$ are the trainable weights and biases of the projector layer respectively. $p$ and $q$ indicate the output of attention devision for premise and hypothesis respectively.

\begin{table*}[ht]
	\small
	\begin{center}
		\resizebox{\textwidth}{!}{
			\begin{tabular}{c|c|c|c|c|c}
				\hline 
				ID & Premise & Hypothesis & Gold & Prediction & Category \\ \hline \hline
				
				\multirow{3}{*}{1} & Six men, two with shirts and four & Seven men, two with shirts and & \multirow{3}{*}{Contradiction} & \multirow{3}{*}{Contradiction} & \multirow{3}{*}{Counting} \\ 
				& without, have taken a break from & four without, have taken a break & & \\ 
				& their work on a building. & from their work on a building. & & \\ \hline
				
				\multirow{3}{*}{2} & two men with shirts and four & Six men, two with shirts and four & \multirow{3}{*}{Entailment} & \multirow{3}{*}{Entailment} & \multirow{3}{*}{Counting} \\ 
				& men without, have taken a break & without, have taken a break from & & \\ 
				& from their work on a building. & their work on a building. & & \\ \hline
				
				\multirow{3}{*}{3} & Six men, two with shirts and four & Six men, four with shirts and two & \multirow{3}{*}{Contradiction} & \multirow{3}{*}{Contradiction} & \multirow{3}{*}{Counting} \\ 
				& without, have taken a break from & without, have taken a break from & & \\ 
				& their work on a building. & their work on a building. & & \\ \hline
				
				\multirow{2}{*}{4} & A man just ordered a book & \multirow{2}{*}{A man ordered a book yesterday.} & \multirow{2}{*}{Neutral} & \multirow{2}{*}{Neutral} & \multirow{2}{*}{Chronology} \\
				& from amazon. & & & \\ \hline
				
				\multirow{2}{*}{5} & A man ordered a book from & \multirow{2}{*}{A man ordered a book yesterday.} & \multirow{2}{*}{Entailment} & \multirow{2}{*}{Entailment} & \multirow{2}{*}{Chronology} \\
				& amazon 30 hours ago. & & & \\ \hline

			\end{tabular}
		}
	\end{center}
	\caption{\label{tab:sample} Examples along their gold labels, ESIM-50 predictions and study categories.}
\end{table*}

\begin{figure*}[ht]
	\centering
	\includegraphics[width=\textwidth]{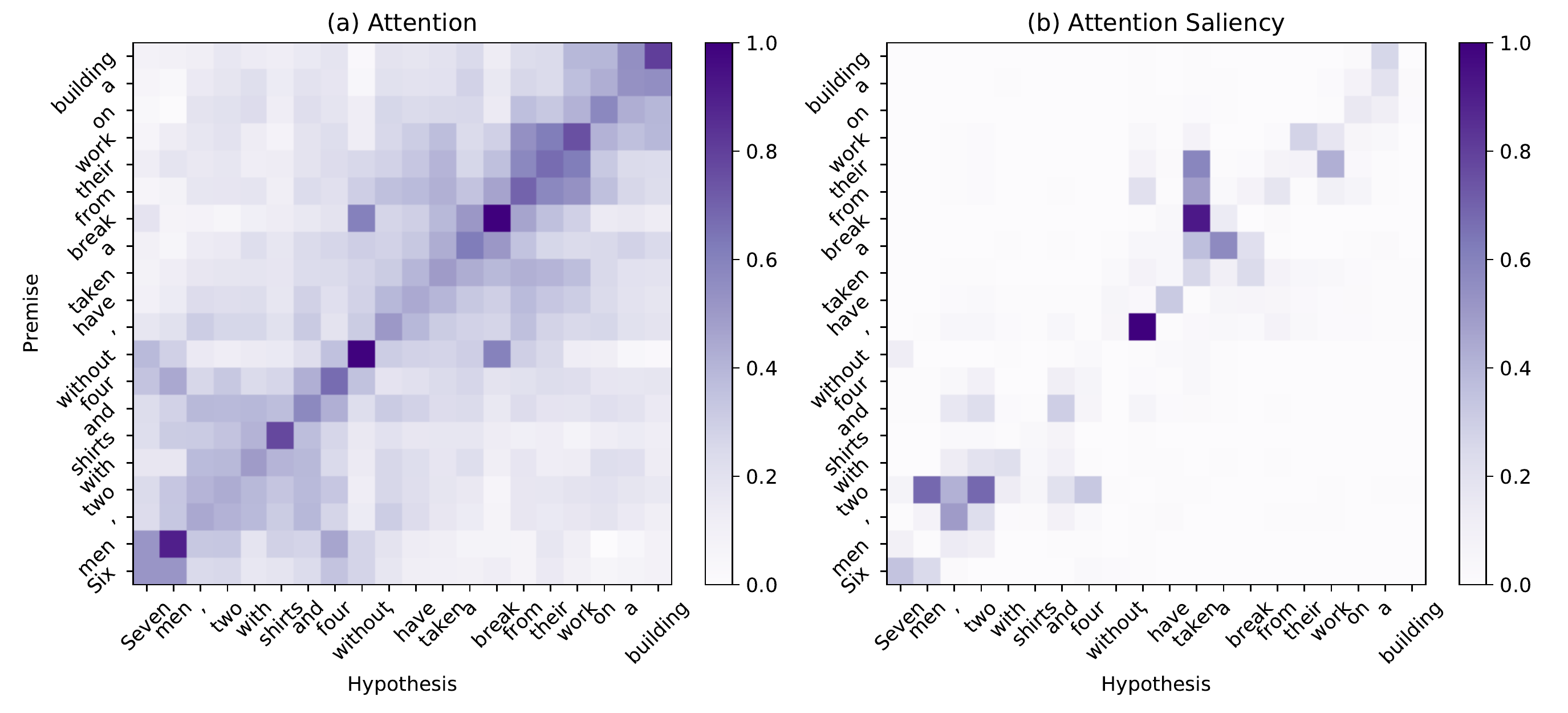}
	\caption{Normalized attention (a) and saliency attention (b) visualizations of Example 1. The gold relationship for this example is Contradiction. ESIM-50 also predicts Contradiction for this example. \label{fig:att_count_seven}}
\end{figure*}

\subsection{Inference}
During this phase, it uses another BiLSTM to aggregate the two sequences of computed matching vectors, $p$ and $q$ from the attention stage (Equations \ref{eq:inf:pout} and \ref{eq:inf:hout}). 	

\begin{equation}
\hat{p} = \textit{BiLSTM}(p)
\label{eq:inf:pout}
\end{equation}

\begin{equation}
\hat{q} = \textit{BiLSTM}(q)
\label{eq:inf:hout}
\end{equation}

\noindent where $\hat{p} \in \mathbb{R}^{n \times 2d}$ and $\hat{q} \in \mathbb{R}^{m \times 2d}$ are the reading sequences of $p$ and $q$ respectively. Finally the concatenation max and average pooling of $\hat{p}$ and $\hat{q}$ are pass through a multilayer perceptron (MLP) classifier that includes a hidden layer with $\emph{tanh}$ activation and $\emph{softmax}$ output layer. The model is trained in an end-to-end manner.

\section{Attention Study}

Here we provide more examples on the NLI task which intend to examine specific behavior in this model. Such examples indicate interesting observation that we can analyze them in the future works. Table 1 shows the list of all example.

\begin{figure*}[ht]
	\centering
	\includegraphics[width=\textwidth]{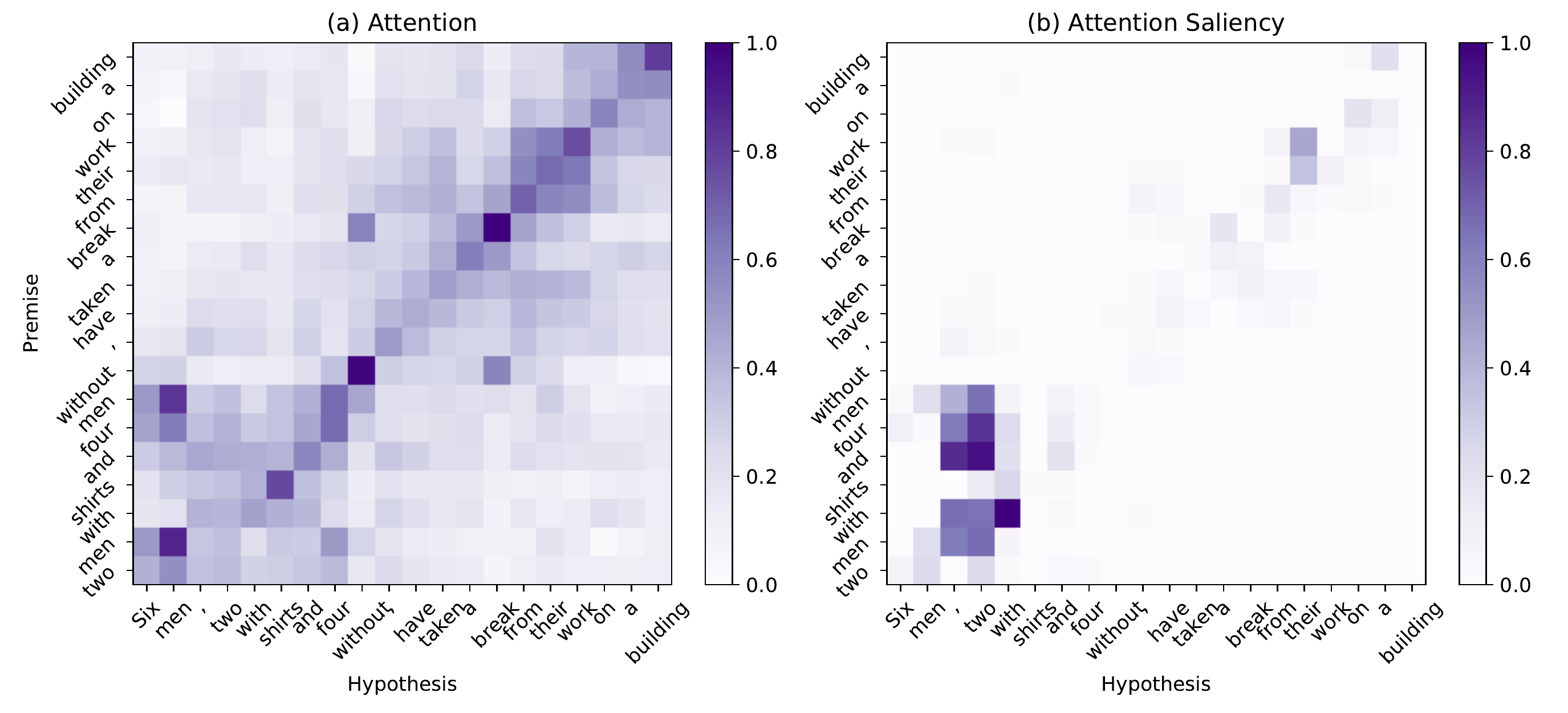}
	\caption{Normalized attention (a) and saliency attention (b) visualizations of Example 2. The gold relationship for this example is Entailment. ESIM-50 also predicts Entailment for this example.  \label{fig:att_count_total}}
\end{figure*}

\begin{figure*}[ht]
	\centering
	\includegraphics[width=\textwidth]{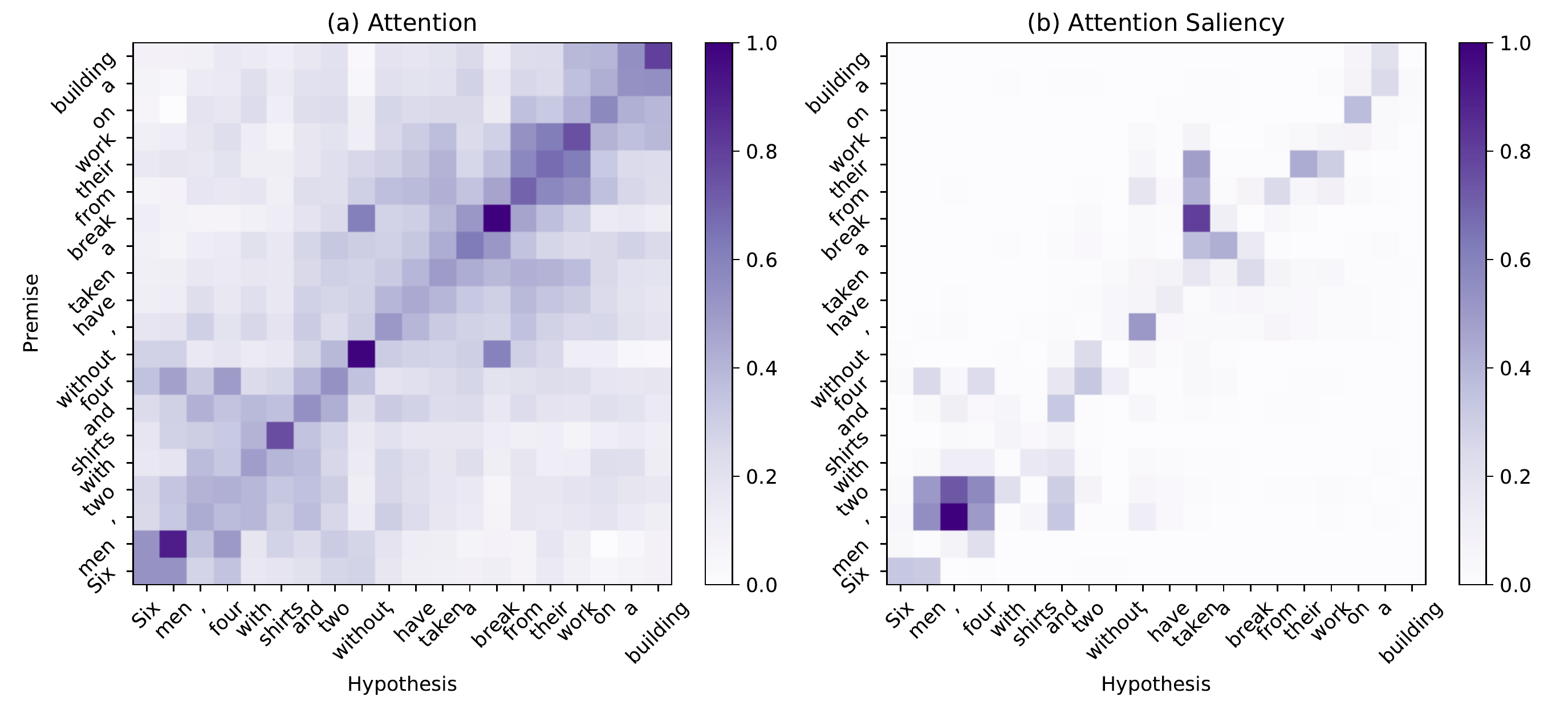}
	\caption{Normalized attention (a) and saliency attention (b) visualizations of Example 3. The gold relationship for this example is Contradiction. ESIM-50 also predicts Contradiction for this example. \label{fig:att_count_sn}}
\end{figure*}

\begin{figure*}[ht]
	\centering
	\includegraphics[width=\textwidth]{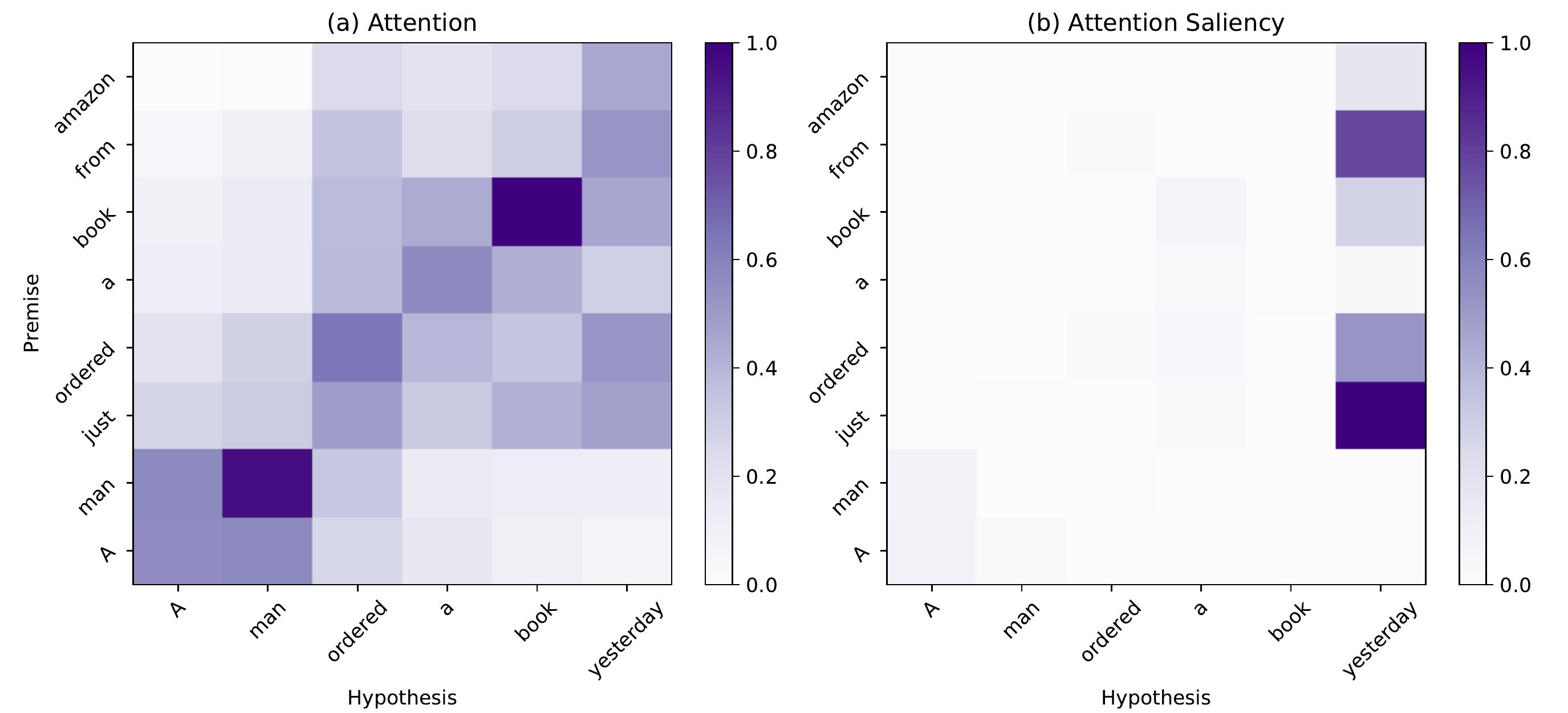}
	\caption{Normalized attention (a) and saliency attention (b) visualizations of Example 4. The gold relationship for this example is Neutral. ESIM-50 also predicts Neutral for this example. \label{fig:att_time_gen}}
\end{figure*}

\begin{figure*}[ht]
	\centering
	\includegraphics[width=\textwidth]{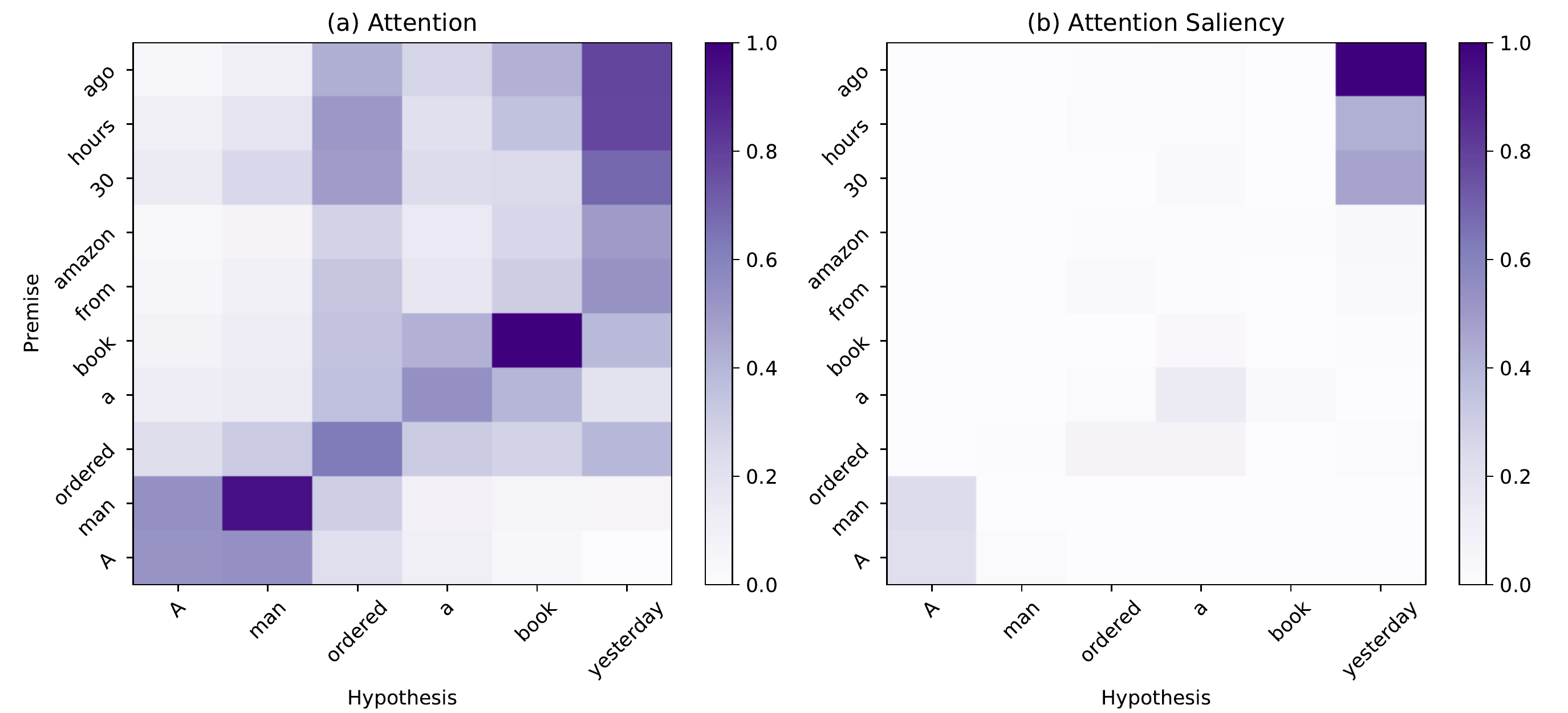}
	\caption{Normalized attention (a) and saliency attention (b) visualizations of Example 5. The gold relationship for this example is Entailment. ESIM-50 also predicts Entailment for this example. \label{fig:att_time_30}}
\end{figure*}

\begin{figure*}[ht]
	\centering
	\includegraphics[width=\textwidth]{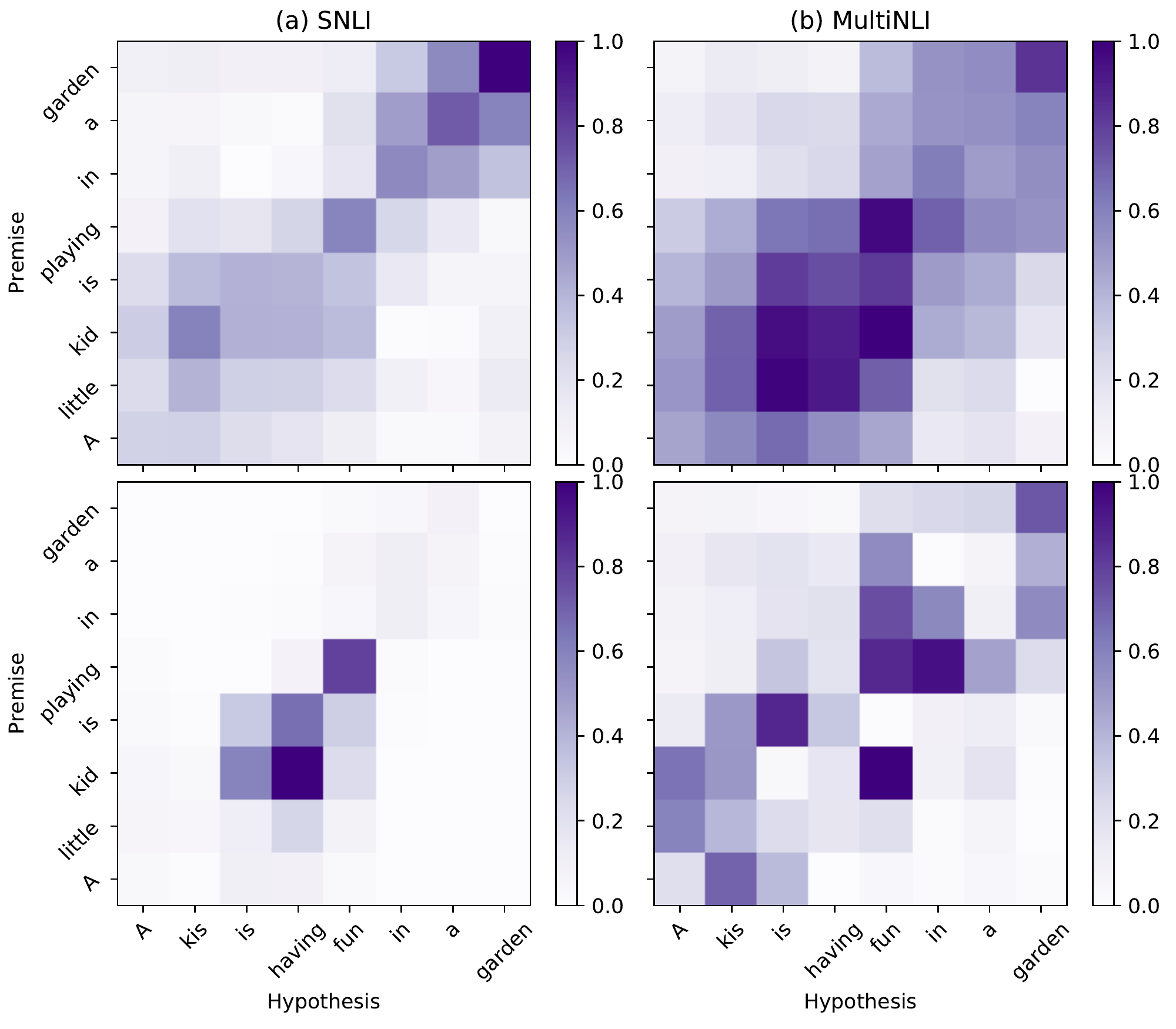}
	\caption{Normalized attention and saliency attention visualizations of an example (h3 in the main paper) for ESIM-50 learned on SNLI (a) and learned on MultiNLI (b). Top plots indicates the attention visualization and bottom ones shows the saliency attention visualization. Both systems correctly predict entailment for this example. \label{fig:att_ed}}
\end{figure*}

\section{LSTM Gating Signal}

\begin{figure*}[ht]
	\centering
	\begin{overpic}[width=\textwidth]{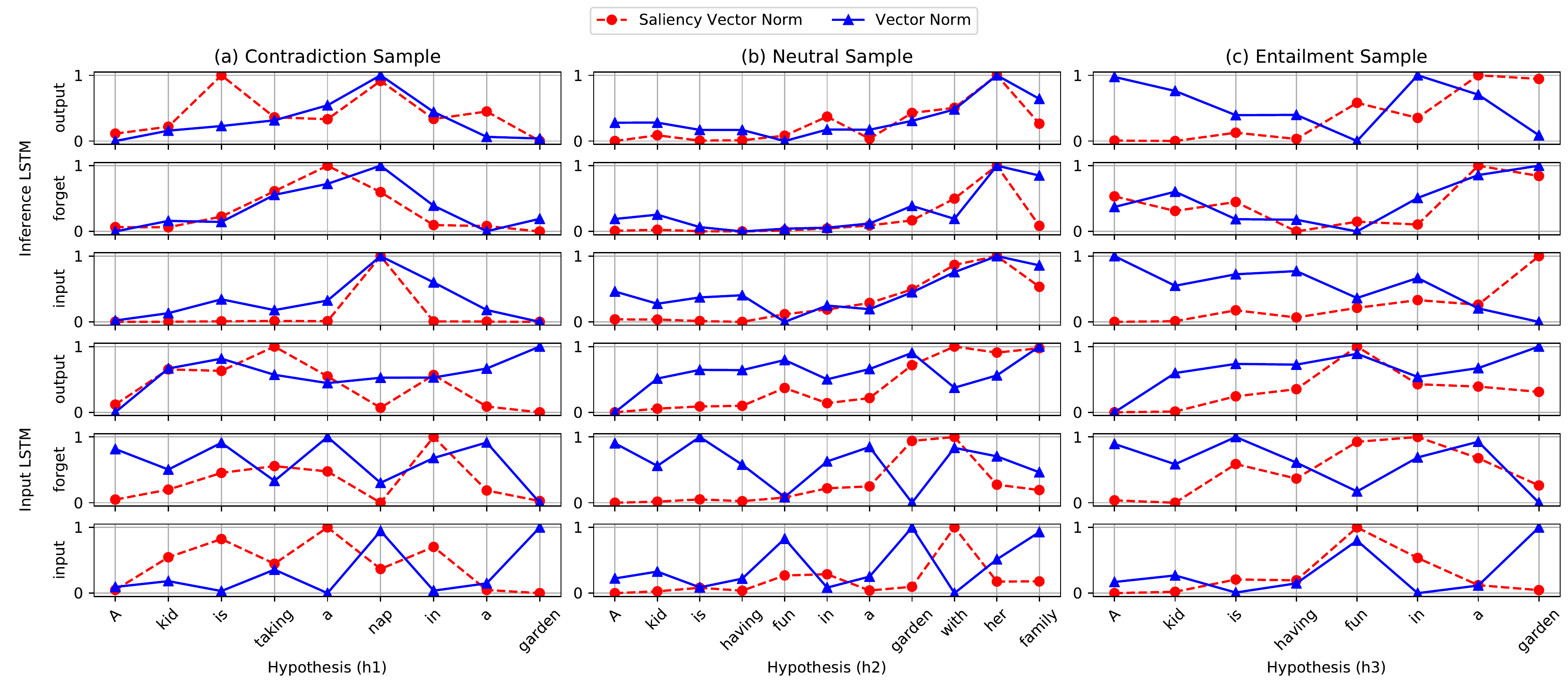}
		\put(0,22.3){\color{black}\rule{\textwidth}{1pt}}
		\put(2.75,5.2){\color{black}\line(0,1){34.1}}
	\end{overpic}
	\caption{Normalized signal and saliency norms for the input and inference LSTMs (backward) for three examples, one for each column. The bottom (top) three rows show the signals of the input (inference) LSTM, where each row shows one of the three gates (input, forget and output). } \label{fig:b_gate_sig}
\end{figure*}

Finally, Figure~\ref{fig:b_gate_sig} depicts the backward LSTM gating signals study.

\end{document}